\documentclass[journal]{IEEEtran}

\usepackage{ifpdf}
\usepackage{xcolor}

\ifCLASSINFOpdf
  
\else
  
\fi

\usepackage[pdftex]{graphicx}
	
  \graphicspath{{./img/}}

\usepackage{amsmath}
\usepackage{amssymb}
\usepackage{multirow}
\usepackage{subfigure}

\usepackage{algorithm}
\usepackage[noend]{algpseudocode}
\usepackage{array}
\usepackage{fixltx2e}
\usepackage{stfloats}
\usepackage{tabularx}
\usepackage{xspace}

\usepackage{amsthm}
	\newtheorem{mydef}{Definition}
	
	\newtheorem{theorem}{Theorem}

\usepackage{fancyhdr}

\fancypagestyle{firstpage}{%
  \lhead{This work has been submitted to the IEEE for possible publication. Copyright may be transferred without notice, after which this version may no longer be accessible.}
}

\begin{document}

\title{Control Design of Autonomous Drone Using Deep Learning Based Image Understanding Techniques}

\author{Seid~Miad~Zandavi,
        Vera~Chung,
				Ali~Anaissi
\thanks{Seid Miad Zandavi, Vera Chung and Ali Anaissi are with the School of Computer Science, The University of Sydney, Camperdown NSW 2006, Australia e-mail: \{miad.zandavi, vera.chung, ali.anaissi\}@sydney.edu.au.}

}


\maketitle
\thispagestyle{firstpage}

\begin{abstract}
This paper presents a new framework to use images as the inputs for the controller to have autonomous flight, considering the noisy indoor environment and uncertainties. A new Proportional-Integral-Derivative-Accelerated (PIDA) control with a derivative filter is proposed to improves drone/quadcopter flight stability within a noisy environment and enables autonomous flight using object and depth detection techniques. The mathematical model is derived from an accurate model with a high level of fidelity by addressing the problems of non-linearity, uncertainties, and coupling. The proposed PIDA controller is tuned by Stochastic Dual Simplex Algorithm (SDSA) to support autonomous flight. The simulation results show that adapting the deep learning-based image understanding techniques (RetinaNet ant colony detection and PSMNet) to the proposed controller can enable the generation and tracking of the desired point in the presence of environmental disturbances.   

\end{abstract}


\IEEEpeerreviewmaketitle

\section{Introduction}

\IEEEPARstart{R}{ecently}, unmanned aerial vehicle (UAV) garners the attention of many researchers working in different application domains such as search and rescue, delivering, and crowdsourcing \cite{kim2019adaptive,koh2012dawn}. UAVs or drones have been developed in many areas, including robotic research, control, path planning, communication, etc. \cite{phung2017enhanced,rajappa2016adaptive,derafa2012super}. Attention to increasing the usability of drones in many commercial/civil applications inspires researchers to make this dynamic system better controllable. In particular, quadcopters are popular drones due to their performance in terms of vertical take-off and landing, simple and stable structure. However, their instability, unstable dynamics, non-linearity, and cross-coupling make this system an interesting under-actuated system. Generally, a quadcopter has six degrees of freedom, although four rotors should control all directions. This causes the cross-coupling between rotation and translational motions. Therefore, the nonlinear dynamics need to be managed by the controller. 

Over recent years, various control algorithms have been developed to deal with the non-linearity of the quadrotor. For example, command-filtered Proportional-Derivative (PD)/ Proportional-Integral-Derivative (PID) control \cite{zuo2010trajectory}, integral predictive control \cite{raffo2010integral} and optimal control \cite{ritz2011quadrocopter,zandavi2018multidisciplinary} have been applied. The Sliding Mode Control (SMC) is another common control algorithm that is used to achieve greater performance in terms of stability due to the influence of modeling errors and external disturbances \cite{derafa2012super,xu2006sliding,besnard2007control}. Note that the chattering effect in the SMC arises in the steady state, where it simulates unmodeled frequencies of the system dynamics.

Of these controllers, PID is preferred due to its simplicity, although it leads to wide overshoot and large settling time \cite{ang2005pid}. Thus, adding a new derivative term (i.e., zero) can decrease the size of the overshoot \cite{jung1996analytic}. This can support better controllability. In addition, this derivative term improves the response in terms of speed and smoothness where limiting overshoot and settling time in an acceptable bound are considered.

Adapting deep learning-based image understanding techniques and the controller opens advancements in drone applications because it improves the capability for autonomous flight without global positioning. Object and depth detection can provide fast, reliable, and integrated information, which is required to reach a target. In this study, the target is a human face, where object detection techniques such as hybrid RentinaNet ant colony detection \cite{zandavi2017novel,lin2017focal} are utilized to recognize the target. To reach the target, depth detention techniques, like PSMNet \cite{chang2018pyramid} are applied to estimate the relative distance to the target.

In this paper, the proposed accelerated PID controller with a derivative filter aims to make an unstable drone/quadcopter track the desired reference with the proper stability. Deep learning and optimization-based image understanding techniques, like object and depth detection, are utilized to make an autonomous drone fly inside buildings and consider uncertainties. The image understanding techniques provide information about a target, such as measuring the relative distance to the recognized target and following a particular threshold. This information is processed to prepare it for the guidance module, which is followed by the control discipline. Thus, the control tracks the desired input, which is generated by the guidance law.

Consequently, the mathematical model of the dynamic system is provided and considers non-linearity, instability, cross-coupling among different modes (i.e., pitch, roll, and yaw), and the uncertain environment. The controller parameters are tuned using the Stochastic Dual Simplex Algorithm (SDSA) optimization algorithm \cite{ZandaviSDSA2019}, which improves the trade-off between exploration and exploitation to achieve better optimal parameters for the proposed controller.

This paper is organized as follows. Section \ref{DroFac_sec2} describes the mathematical model of the dynamic system.The proposed controller is explained and introduced in Section \ref{DroFac_sec3}. Stability analysis is presented in Section \ref{DroFac_sec4} and SDSA is briefly explained in Section \ref{DroFac_sec5}. Guidance law is then introduced in Section \ref{DroFac_sec6} and deep learning-based image understanding techniques are introduced in Section \ref{DroFac_sec7}. The numerical results and discussion are presented in Section \ref{DroFac_sec8}. Finally, this paper ends with a conclusion. 


\section{Dynamic Model}\label{DroFac_sec2}
\begin{figure}[h]
    \centering
    \includegraphics[scale = 0.7]{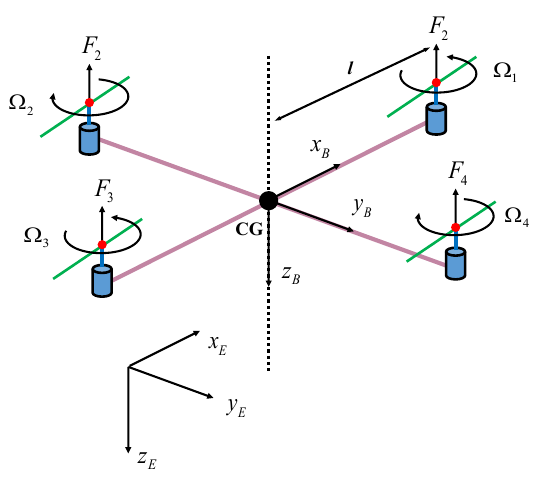}
    \caption{Earth Fixed and Body Fixed coordinate systems}
    \label{fig_Control1}
\end{figure}

The mathematical model of a system can be used as the first step to study its performance. In this regard, the quadcopter studied in this research is modeled in Fig. \ref{fig_Control1}, considering earth-centered inertia (ECI) and body frame. Thus, $X_E = [x_E , y_E , z_E]^T$ and $X_B = [x_B , y_B , z_B]^T$ are defined as transformational motions from inertia frame to body frame due to having an accurate dynamic model.


The attitude of the quadcopter is formulated based on the Eular angles roll, pitch, and yaw, which are rotated from the x-axis, y-axis and z-axis, respectively. Thus, the Eular angles are $\Theta = [\phi, \theta, \psi]^T$, and the angular velocity in the body frame is $\dot{\Theta} = [\dot{\phi}, \dot{\theta}, \dot{\psi}]^T$. In this sense, the angular velocity in inertia ($\omega = [p, q, r]^T$) is formulated as follows:

\begin{equation}
\label{Control_eq1}
	\omega =  \left[ {\begin{array}{ccc}
									1 & 0 & -\sin(\theta) \\
									0 & \cos(\phi) & \cos(\theta) \sin(\phi) \\
									0 & -\sin(\phi) & \cos(\theta) \cos(\phi)  \end{array}} \right] \cdot \dot{\Theta}
\end{equation}

Total torques are caused by three segments: thrust forces ($\tau$), body gyroscopic torque (${\tau}_b$) and aerodynamic friction (${\tau}_a$). In addition, each component of the torque vector ($\tau = [{\tau}_{\phi},{\tau}_{\theta},{\tau}_{\psi}]^T$), corresponding to a rotation in the roll, pitch, and yaw axis, can be determined by Eqs (\ref{Control_eq2})$-$(\ref{Control_eq4}):

\begin{equation}
\label{Control_eq2}
{\tau}_{\phi} = l (F_2 - F_4)
\end{equation}

\begin{equation}
\label{Control_eq3}
{\tau}_{\theta} = l (F_3 - F_1)
\end{equation}

\begin{equation}
\label{Control_eq4}
{\tau}_{\psi} = c (F_2 - F_1 + F_4 - F_3)
\end{equation}
where \textit{l} is the distance between the center of motor and the center of mass, and \textit{c} is the force to torque coefficient. As assumed, the quadcopter is a rigid body and symmetrical dynamics apply, from which the torque can be calculated by following equation: 

\begin{equation}
\label{Control_eq5}
{\tau} = I \dot{\omega} + \Omega (I \omega)
\end{equation}

where \textit{l} is the distance between the center of motor to center of mass, and \textit{c} is the force to torque coefficient. As assumed, the quadcopter is a rigid body and symmetrical dynamics, from which the following equation can calculate the torque: 

\begin{equation}
\label{Control_eq6}
{\Omega} = \left[ {\begin{array}{ccc}
									0 & -r & q \\
									r &  0 & -p \\
									-q & p & 0 \end{array}} \right]
\end{equation}

In this system, the main control inputs are correlated to the torque ($\tau = [{\tau}_{\phi},{\tau}_{\theta},{\tau}_{\psi}]^T$) caused by thrust forces, body gyroscopic effects, propeller gyroscopic effects and aerodynamic friction. Gyroscopic effects and aerodynamic friction are considered external disturbances for the control. Thus, control inputs are determined as Eq (\ref{Control_eq7}).

\begin{equation}
\label{Control_eq7}
\left[ {\begin{array}{c}
u_{\phi} \\ u_{\theta}\\ u_{\psi} \\ u_{T} 
\end{array}} \right]
=
\left[ {\begin{array}{c}
\tau_{\phi} \\ \tau_{\theta} \\ \tau_{\psi} \\ \tau_{T} 
\end{array}} \right]
=
\left[ {\begin{array}{cccc}
0 & l & 0 & -l \\
-l & 0 & l & 0 \\
-c & c & -c & c \\
1 & 1 & 1 & 1
\end{array}} \right]
\left[ {\begin{array}{c}
F_1 \\ F_2 \\ F_3 \\ F_4 
\end{array}} \right]
\end{equation}
where $\tau_{T}$ is the lift force and $u_T$ corresponds to the total thrust acting on the four propellers, where $u_{\phi}$, $u_{\theta}$ and $u_{\psi}$ represent the roll, pitch, and yaw, respectively. The drone’s altitude can be controlled by lift force ($u_T$), which is equal to quadcopter weight. The dynamic equations of the quadcopter are formulated based on the Newton-Euler method \cite{zipfel2007modeling}. The six degree of freedom (6-DOF) motion equations are stated by Eqs (\ref{Control_eq8})$-$(\ref{Control_eq13}).

\begin{equation}
\label{Control_eq8}
\dot{u} = rv -qw -g \sin(\theta)
\end{equation}

\begin{equation}
\label{Control_eq9}
\dot{v} = pw -ru +g  \sin(\phi) \cos(\theta)
\end{equation}

\begin{equation}
\label{Control_eq10}
\dot{w} = qu -pv + g \cos(\theta) \cos(\phi) - \frac{1}{m} u_T
\end{equation}

\begin{equation}
\label{Control_eq11}
\dot{p} = \frac{1}{I_{xx}} \left[ (I_{yy}-I_{zz})qr + u_{\phi} + d_{\phi} \right]
\end{equation}

\begin{equation}
\label{Control_eq12}
\dot{q} = \frac{1}{I_{yy}} \left[ (I_{zz}-I_{xx})pr + u_{\theta} + d_{\theta} \right]
\end{equation}

\begin{equation}
\label{Control_eq13}
\dot{r} = \frac{1}{I_{zz}} \left[ (I_{xx}-I_{yy})pq + u_{\psi} + d_{\psi} \right]
\end{equation}
where $d = [d_{\phi},d_{\theta},d_{\psi}]^T$ is the angular acceleration disturbance corresponded to propeller angular speed, and these acceleration disturbances are modeled by Eq (\ref{Control_eq14}).

\begin{equation}
\label{Control_eq14}
 d = \left[ {\begin{array}{c}
									+  qI_{m}  \Omega_{r} \\ -pI_{m} \Omega_{r} \\ 0 
									\end{array}} \right]
\end{equation}
where $\Omega_r = \sum_{i=1}^{4} (-1)^{i} \Omega_i $ is the overall residual propeller angular speed, and $\Omega_i$ is the angular velocity of each rotor. $I_{m}$ is the rotor moment of inertia around the axis of rotation. Hence, the dynamics equations of the system can be summarized as follows: 

\begin{equation}
\label{Control_eq15}
{\begin{array}{c}
\dot{x}(t) = A(x) + B(x)u(t) + d  \\
   y(t) = C(x) +D(x)u(t)
	\end{array}}
\end{equation}
where $x = [\phi, \theta, \psi, p, q, r, w]^T$ and $y = [y_1,y_2,y_3,y_4]^T$ are the states and measurable outputs, respectively. $u = [u_1,u_2,u_3,u_4]^T$ is the control and $d$ is the disturbance. $A$, $B$, $C$, and $D$ are the nonlinear functions regarding dynamic equations of the system. 

The control design is considered to minimize the error for tracking the desired command (see Eq (\ref{Control_eq16})).

\begin{equation}
\label{Control_eq16}
\lim_{t\to\infty} |e(t)| = \varepsilon
\end{equation}
where $e(t) = r(t)-y(t)$ is the difference between reference inputs and the system’s measurable outputs and $\varepsilon$ is the small positive value.


\section{Proposed PIDA Controller} \label{DroFac_sec3}
The PID control is used in many engineering applications because of its simplicity. Note that PID cannot function well when wide overshoot and large settling time occur in the system. This issue can be addressed by a modifying the PID controller by adding an additional zero known as PIDA. It is employed to achieve a faster and smoother response for higher-order systems and retains both overshoot and settling time within an acceptable limit. Additionally, the proposed linear control is able to control the nonlinear system. In this approach, the dynamic airframe is linearized about the equilibrium point. The linearization of the model is given by Eq (\ref{DroFac_eq17}).
\begin{equation}
\label{DroFac_eq17}
\Delta \dot{X} = J_X \Delta X + J_U \Delta U 
\end{equation}
where $J_X$ and $J_U$ are the Jacobian transformation of the nonlinear model about the equilibrium point ($X_{eq} = [\phi_0,\theta_0,\psi_0,p_0,q_0,r_0,w_0]^T$). Note that the equilibrium point can be calculated by solving $\dot{X} = AX = 0$. Any solution can be the equilibrium point because of null space if $det(A)$ is equal to zero. 

In this regard, a Multi-Input Multi-Output (MIMO) control system design follows the desire command in altitude and attitude channels. A MIMO tracking controller can not only stabilize the system, but also make it follow a reference input. Thus, the linear system is given as follows: 

\begin{equation}
\label{DroFac_eq18}
{\begin{array}{c}
\dot{X} = AX + BU + D_d\\
   Y = CX
	\end{array}}
\end{equation}
where $Y$ is the outputs that follow the reference inputs and $D_d = [0, 0, 0, d^T, 0]^T$ is the angular disturbance. In this approach, the integral state is defined as follows: 

\begin{equation}
\label{DroFac_eq19}
\dot{X}_N = R-Y = R-CX 
\end{equation}

According to Eq (\ref{DroFac_eq19}), the new state space of the system is formulated in Eq (\ref{DroFac_eq22}). It is obvious that the system can follow the reference inputs if the designed controller proves the stability of the system. 

\begin{equation}
\label{DroFac_eq22}
{\begin{array}{c}
\left[ {\begin{array}{c}
\dot{X} \\
   \dot{X}_N 
	\end{array}}\right]= 
	\left[ {\begin{array}{cc}
	A & 0 \\
	-C & 0
	\end{array}}\right]
	\left[ {\begin{array}{c}
		{X} \\
		{X}_N 
	\end{array}}\right]+\left[ {\begin{array}{c}
		B \\
		\Phi
	\end{array}}\right]U +
	\\
	\left[ {\begin{array}{c}
\Phi \\
   I 
	\end{array}}\right] R + \left[ {\begin{array}{c} I \\ \Phi \end{array}}\right]D_d  \\
	\\
	Y = \left[{\begin{array}{cc}
	C & 0 \end{array}}\right] \left[{\begin{array}{c}
	X \\
	X_N
	\end{array}}\right]	
	\end{array}}
\end{equation}
where $\Phi$ is a zero matrix. 

Regarding the acceleration disturbance in the system, the general form of the proposed controller in the time series is given in Eq (\ref{DroFac_eq23}). 

\begin{equation}
\label{DroFac_eq23}
u(t) =  k_p e(t) + k_i \int{e(t) dt}+ k_d \dot{e}(t) + k_a \ddot{e}(t)
\end{equation}
where $kp$, $k_i$, $k_d$, and $k_a$ are the gain of the proposed controller. Then, the MIMO controller is generated by:
\begin{equation}
\label{DroFac_eq24}
U(s) = \left[ k_p + \frac{k_i}{s} + k_d s + k_a s^2 \right] E(s)
\end{equation}

As seen in Eq (\ref{DroFac_eq24}), the derivative term is not efficient in a high-frequency domain. This term can affect the performance of the whole system in a noisy environment. Adding derivative filter is proposed to address this issue and thus, the proposed control is modeled as follows: 

\begin{equation}
\label{DroFac_eq25}
U(s) = \left[ k_p + \frac{k_i}{s} + k_d \times s L(s) + k_a \times s L(s) \times s L(s) \right] E(s)
\end{equation}
where $L(s)$ is the optimal derivative filter which is formulated as follows:

\begin{equation}
\label{DroFac_eq20}
L(s) = \frac{N/T}{(N/T) \frac{1}{s}+ 1}
\end{equation}
where $N$ and $T$ are the order of the filter and time constant, respectively. Based on Eq (\ref{DroFac_eq20}), the transfer function of optimal derivative filter can be simplified as follows: 

\begin{equation}
\label{DroFac_eq21}
L(s) = \frac{1}{1 + T_f s}
\end{equation}
where $T_f = T/N$ is the time constant of the optimal derivative filter. Hence, the controller and filter’s parameters can be found by SDSA to minimize the objective function given by Eq (\ref{DroFac_eq222}).

\begin{equation}
\label{DroFac_eq222}
f_{obj} = (M_{os}-M_s)^2 - (t_s - t_s)^2
\end{equation}
where $M_{os}$ is the desired maximum overshoot, which is set to $5$ percent; $t_s$, the desired settling time for the system, is $2~sec$. $M_s$ and $t_s$ are the overshoot and settling time for each set of designed controllers. Before the simulation result is presented, the stability analysis of the system is introduced in Section \ref{DroFac_sec4}.


\section{Stability Analysis of the Proposed PIDA} \label{DroFac_sec4}
In this section, the stability of a system considering the proposed controller is investigated. The following definitions are needed.

\begin{mydef}
\label{def1}
“Asymptotically stable” is a system around its equilibrium point if it meets the following conditions:

\begin{enumerate}
	\item Given any $\epsilon > 0$, $\exists$ ${\delta}_{1} > 0$ such that if $\|x(t_0)\| < \delta_1$, then $\|x(t)\| < \epsilon$, $\forall$ $t$ \quad > \quad $t_0$ 
	
	\item $\exists$ $\delta_2 > 0$ such that if $\|x(t_0)\| < \delta_2$, then $x(t) \to 0$ as $t \to \infty$
	
\end{enumerate}

\end{mydef}

\begin{theorem}
$[V(x) = x^T P x, \quad x \in \mathbb{R}^n]$ is a positive definite function if and only if all the eigenvalues of $P$ are positive. 
\end{theorem}

Since $P$ is symmetric, it can be diagonalized by an orthogonal matrix so $P=U^T D U$ with $U^T U = I$ and $D$ diagonal. Then, if $y = Ux$,
\begin{equation}
\begin{split}
    V(x) &= x^T P X \\
		&= x^T U^T D U x \\ 
    &= y^T D y \\
    &= \sum {\lambda}_i |{y_i}|^2 \\
\end{split}
\end{equation}
	
Thus,
\begin{equation}
	V(x) > 0 \quad \forall x \neq 0  \iff \lambda_i > 0, \quad \forall i
\end{equation}

\begin{mydef}
\label{def2}
A matrix $P$ is a positive definite if it satisfies $x^T P x > 0 \quad \forall x \neq 0$.
\end{mydef}

Therefore, any positive definite matrix follows the inequality in Eq (\ref{DroFac_eq27}).

\begin{equation}
\label{DroFac_eq27}
\lambda_{min} P \|x\|^2 \leq V(x) \leq \lambda_{max} P \|x\|^2
\end{equation}
 
\begin{mydef}
\label{def3}
($V$) is a positive definite function as a candidate Lyapunov function if ($\dot{V}$) has derivative, and it is negative semi-definite function. 
\end{mydef}

\begin{theorem}
\label{theorem1}
If the candidate Lyapunov function (i.e., $V(x) = x^T P x, \quad P>0$) exists for the dynamic system, there is a stable equilibrium point.
\end{theorem}

According to Theorem \ref{theorem1} and the dynamic system model, the system in the form of Lyapunov function is as follows: 

\begin{equation}
\begin{split}
    \dot{V}(x) & = \dot{x}^TPx + x^TP\dot{x}\\
		& = x^T A^T P x + x^T P A x \\
		& = x^T(A^T P + P A)x \\
		& = -x^T Q x 
\end{split}
\end{equation}
where the new notation (see Eq (\ref{DroFac_eq29})) is introduced to simplify the calculation. It is noted that $Q$ is a symmetric matrix. According to Definition \ref{def3}, $V$ is a Lyapunov function if $Q$ is positive definite (i.e., $Q>0$). Thus, there is a stable equilibrium point which shows the stability of the system around the equilibrium (see Theorem \ref{theorem1}). 

\begin{equation}
\label{DroFac_eq29}
		A^T P + P A = -Q
\end{equation}

The relationship between $Q$ and $P$ shows that the solution of Eq (\ref{DroFac_eq29}), called a Lyapunov equation, proves the stability of the system for picking $Q > 0$ if $P$ is a positive definite solution. Thus, there is a unique positive definite solution if all the eigenvalues of $A$ are in the left half-plane. A noisy environment causes the movement of eigenvalues to the right half-plane. Therefore, the system dynamics can intensify instability. This issue raises the cross-coupling among different modes such as roll, pitch, and yaw rate, which are caused by the four rotors. Thus, the derivative term of the proposed controller plays an essential role in maintaining stability. The numerical results show that all eigenvalues of the quadcopter with considering the proposed controller with uncertainties in the environment are in the left half-plane, which proves that the dynamic system is stable with uncertainties.


\section{Stochastic Dual Simplex Algorithm} \label{DroFac_sec5}

The heuristic optimization algorithm (i.e., SDSA) is used to find the best tuned parameters for the proposed controller. SDSA is a new version of the Nelder-Mead simplex algorithm \cite{rao2009engineering}, executing three different operators, such as reflection, expansion, and contraction. These operators reshape the dual simplex and move it toward the maximum-likelihood regions of the promising area. Each simplex follows the standard rules of simplex, from which the transformed vertices of the general simplex approach are formulated, as in Eqs (\ref{DroFac_eq30})-(\ref{DroFac_eq32}). 

\begin{equation}
\label{DroFac_eq30}
\textbf{x}_r = (1+\alpha)\bar{\textbf{x}}_0 - \alpha \textbf{x}_h , \quad \alpha > 0
\end{equation}

\begin{equation}
\label{DroFac_eq31}
\textbf{x}_e = \gamma \textbf{x}_r + (1-\gamma)\bar{\textbf{x}}_0 ,   \quad \gamma > 1
\end{equation}

\begin{equation}
\label{DroFac_eq32}
\textbf{x}_c = \beta \textbf{x}_h + (1-\beta)\bar{\textbf{x}}_0 , \quad  0 \leq \beta \leq 1
\end{equation}
where $\alpha$, $\gamma$, and $\beta$ are reflection, expansion, and contraction coefficients, respectively. During these transformations, the centroid of all vertices excluding the worst point ($\textbf{x}_h$) is $\bar{\textbf{x}}_0$. 

In addition to the movement of dual simplex, a new definition of reflection points is applied to improve the diversity and decrease the probability of a local minimum. Therefore, during the \textit{i}-th iteration, the worst vertices of simplexes in the search space are replaced by normal distribution directions, which are modeled in Eq (\ref{DroFac_eq33}). 

\begin{equation}\label{DroFac_eq33}
\overset{*}{\textbf{x}}_{h_s}^{(i)} = \textbf{x}_{h_s}^{(i)} + g^{(i)} {\bar{\textbf{x}}_0}^{(i)}
\end{equation}
where $\overset{*}{\textbf{x}}_{h_s}^{(i)}$ is the new reflected point computed by the worst point of each simplex (${\textbf{x}}_{h_s}^{(i)}$), and $g^{(i)}$ is the normal distribution of the sampled solution in \textit{i}-th iteration and $s$-th simplex. The centroid of all simplexes and the probability density function of the normally distributed simplexes are then expressed in Eq (\ref{DroFac_eq34}) and  Eq (\ref{DroFac_eq35}).

\begin{equation}\label{DroFac_eq34}
\bar{\textbf{x}}_0^{(i)} = \sum_{s=1}^{n_s} {\bar{\textbf{x}}_{0_s}^{(i)}}
\end{equation}

\begin{equation}\label{DroFac_eq35}
g(\textbf{x}_h|\Sigma) = \frac{1}{\sqrt{2\pi|\Sigma|}}.exp({-\frac{(\textbf{x}_h-\bar{\textbf{x}}_0)^T{\Sigma}^{-1}(\textbf{x}_h-\bar{\textbf{x}}_0)}{2}})
\end{equation}
where $n_s$ and $\Sigma$ are the number of simplexes and covariance matrix of simplexes, respectively.

\par
Reflection makes an action regarding reflect the worst point, called high, over the centroid $\bar{\textbf{x}}_0$. In this approach, simplex operators utilize the expansion operation to expand the simplex in the reflection direction if the reflected point is better than other spots. Nevertheless, the reflection output is at least better than the worst; the algorithm repeats the reflection operation with the new worst point \cite{ZandaviSDSA2019,rao2009engineering}. The contraction is another operation that contracts the simplex, while the worst point has the same value as the reflected point. The SDSA pseudo-code is presented in Algorithm \ref{SDSA}, and the tuned parameters of SDSA, chosen based on \cite{ZandaviSDSA2019}, are listed in Table \ref{DroFac_table1}.

\begin{algorithm}
\caption{Stochastic Dual Simplex Algorithm (SDSA)}\label{SDSA}
\begin{algorithmic}

\State \textbf{Initialization}

\State \quad{\textit{set}} $\gets$ \big[$\textit{a}_{max}$,$\alpha_{max}$, $\gamma_{max}$, $\beta_{max}$, $i_{max}$ \big]

\State \quad{$\textbf{x}_0$} $\gets$ \textit{random}

\State \quad{Generate initial $simplexes$}

\State \textbf{Repeat}

\State \quad{Compute Objective Function $(F)$}
\State \quad{$\textbf{x}_h$} $\gets$ $\textbf{x}_{worst}$

\State \quad\quad{\textbf{while}\;({$\exists \; x_i$})}:
\State \quad\quad\quad\quad{$reflection$}
\State \quad\quad\quad\quad{$expansion$}
\State \quad\quad\quad\quad{$contraction$}
\State \quad\quad{\textbf{end}}

\State \quad{$\textbf{x}_h$} $\gets$ $\overset{*}{\textbf{x}_h}$

\State \quad{Update $simplexes$}

\State \textbf{Until} Stop condition satisfied.

\end{algorithmic}
\end{algorithm}

\begin{table}[h]
\centering
\caption{Tuned parameters of SDSA}
\label{DroFac_table1}
\begin{tabular}{l ccc}
\hline
\textbf{Parameters} &&& \textbf{Value}  \\ \hline

${a}_{max}$ &&& $10.5907$ \\
${\alpha}_{max}$ &&& $9.7323$ \\
${\gamma}_{max}$ &&& $9.9185$ \\
${\beta}_{max}$ &&& $0.4679$ \\
${\textit{i}}_{max}$ &&& $979$ \\

\hline

\end{tabular}
\end{table}

\section{Guidance Law} \label{DroFac_sec6}
Guidance law is a process to generate commands for a pursuer to track its target. An interesting guidance law is proportional navigation (PN) which considers the rotation rate of line of sight (LOS) \cite{siouris2004missile}. The general concept of PN is to vary the lateral acceleration of the pursuer in proportion with the rotation rate of LOS. In this study, pure proportional navigation (PPN) guidance law is applied to the quadcopter for tracking and reaching the target. In PPN, the desired acceleration command (proportional to LOS turning rate) is applied normal to the velocity vector of the pursuer. Therefore, PPN is modeled as follows:

\begin{equation}\label{DroFac_eq36}
a_c = N \Omega_{LOS} \times (V_M-V_T)
\end{equation} 
where $N$ is the constant number, named navigation constant ($N = 1$). $\Omega_{LOS}$, $V_M$, and $V_T$ are the angular velocity of LOS, pursuer velocity (drone velocity), and the target velocity, respectively. The angular velocity of LOS is formulated in Eq (\ref{DroFac_eq37}).

\begin{equation}\label{DroFac_eq37}
\Omega_{LOS} = \frac{(V_M-V_T) \times R}{|R|^2}
\end{equation} 
where $R$ is the relative distance between the pursuer and the target. $R$ is estimated by image depth detection. 
In this study, the drone is simulated to fly inside the indoor environment of a building, where positioning systems such GPS are unable to operate. In this regard, object tracking and depth detection are employed to estimate the relative distance between the drone and target. The target is set as a building mockett, considering the safe distance from the target. The following section explains the object and depth detection approaches.

\section{Deep Learning Based Image Understanding \\Techniques}\label{DroFac_sec7}

This section proposes a framework to adapt deep learning-based image understanding techniques using object and depth detection for generating and translating the target to the drone. Object and depth detection play an important role in autonomous drones without positioning systems. Note that the indoor environment is furnished with many objects and a particular object (a building mockett) is considered a target for the drone to reach. 

Object detection is an indispensable aspect of positioning in autonomous drone flight inside a building because it is the target point to obtain the relative distance estimation ($R$). While the relative distance is estimated through depth detection, the guidance law (PPN) is used to issue such a command to the controller to track the target. In this study, RetinaNet ant colony detection \cite{lin2017focal} is applied in the detection module of the proposed autonomous system. 

\begin{figure*}
    \centering
    \includegraphics[scale = 0.65]{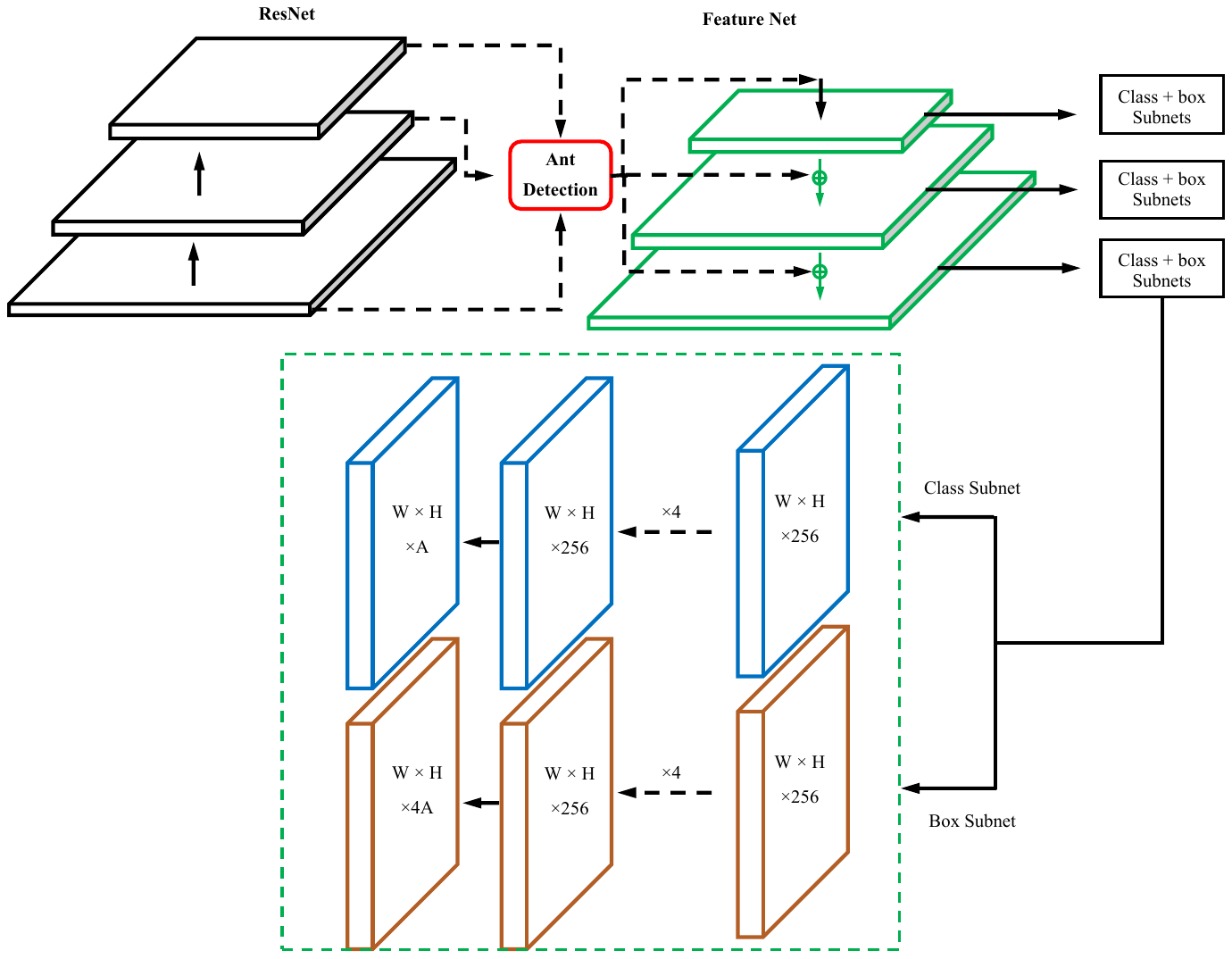}
    \caption{The architectuer of RetinaNet Ant Colony Detection}
    \label{fig_AIFace}
\end{figure*}

Ant Colony Detection (ACD) \cite{zandavi2017novel} utilizes a novel multi-region feature selection method that defines histogram values of basic areas and random areas (MRH), and is combined with Continuous Ant Colony Filter (CACF) as a heuristic filter \cite{nobahari2016simplex,zandavi2019state} detection to represent the original target. In this paper, the ant colony detection is combined with the popular one-stage RetinaNet \cite{lin2017focal} using five modification steps. First, classification and regression are applied for detection. Second, the intersection over union (IoU) loss function is used for regression. Third, reconsideration is augmented regarding data-anchor-sampling for training. Forth, there is a robust classification utilizing the max-out operation and fifth, the multi-scale testing strategy is applied for inference. Consequently, the proposed method achieved efficient performance in the recognition of the target. The RetinaNet ant colony detection’s architecture is presented in Fig. \ref{fig_AIFace}.

The proposed method uses a RentinaNet \cite{lin2017focal} object detector baseline considering the particular focal loss function to address extreme class imbalance encountered during training. The focal loss function is formulated in Eq (\ref{DroFac_eq38} and \ref{DroFac_eq39}).

\begin{equation}\label{DroFac_eq38}
FL(p_t) = -\nu_t(1-p_t)^\tau \log(p_t)
\end{equation}

and

\begin{equation} 
\label{DroFac_eq39}
\begin{split}
p_t = \Big\{ {\begin{array}{cc}
p & \text{if} \quad y = 1 \\
1-p & \text{otherwise} \\
\end{array}} 
\end{split}
\end{equation}
where ${y} \in \{ {\scriptstyle  \pm} {1} \}$ defines the ground-truth class, $p \in [0,1]$ is the model’s estimated probability for the class labeled $y = 1$, and $\nu_t$ and $\tau$ are the balanced and focusing parameters, respectively. 

The object detection component includes the classification and regression sub-tasks. For the regression task, the UnitBox \cite{yu2016unitbox} is applied to minimize the differences between the predictions and ground-truth through IoU, rather than using the smooth $L1$ \cite{girshick2015fast} as the common loss function for the regression. Thus, the IoU regression loss function is modeled in Eq (\ref{DroFac_eq40}).

\begin{equation}\label{DroFac_eq40}
L_{IoU} = - \ln {\frac{Intersection(B_p,B_{gt})}{Union(B_p,B_{gt})}}
\end{equation}
where $B_p = (x_1,y_1,x_2,y_2)$ and $B_{gt} = (x^*_1,y^*_1,x^*_2,y^*_2)$ are the predicted and the ground-truth bounding boxes, respectively. Consequently, IoU calculates the similarity between the truth and predicted bounding boxes. The minimization of this similarity improves algorithm performance in the regression sub-task. 

In addition to the regression loss function, two-step classification (STC) and selective two-step regression (STR) are employed in the selective refinement network \cite{chi2019selective}. In this regard, STC and STR conduct a two-step classification on three low-level detection layers and two-step regression on three high-level detection layers, respectively. These loss functions are formulated in Eq (\ref{DroFac_eq41})$-$(\ref{DroFac_eq42}).

\begin{equation}\label{DroFac_eq41}
\begin{split}
L_{STC}(p_i,q_i) & = \frac{1}{N_{s_1}} \sum_{i \in \Upsilon} L_{FL}(p_i,{l^*}_i) \\
& +\frac{1}{N_{s_2}} \sum_{i \in \Gamma} L_{FL}(q_i,{l^*}_i) \\
\end{split}
\end{equation}

and 

\begin{equation}\label{DroFac_eq42}
\begin{split}
L_{STR}(r_i,t_i) & = \frac{1}{N_{s_1}} \sum_{i \in \Psi} [{l^*}_i=1] L_{IoU}(r_i,{g^*}_i) \\
& + \frac{1}{N_{s_2}} \sum_{i \in \Gamma} [{l^*}_i=1] L_{IoU}(t_i,{g^*}_i) \\
\end{split}
\end{equation}
where $i$, $p_i$/$q_i$, and $r_i$/$t_i$ are the anchor index, first/second step of predicted classification and regression, respectively, and $l^*_i$/$g^*_i$ are the class/location ground-truth. The number of positive anchors is $N_{s_1}$/$N_{s_2}$ for first/second step, and $\Upsilon/\Psi$ and $\Gamma$ are the collection of classification/regression samples for the first and second steps. $L_{FR}$ is also the sigmoid focal loss function, which is formulated in Eq (\ref{DroFac_eq38}). 

\begin{figure*}
    \centering
    \includegraphics[scale = 0.70]{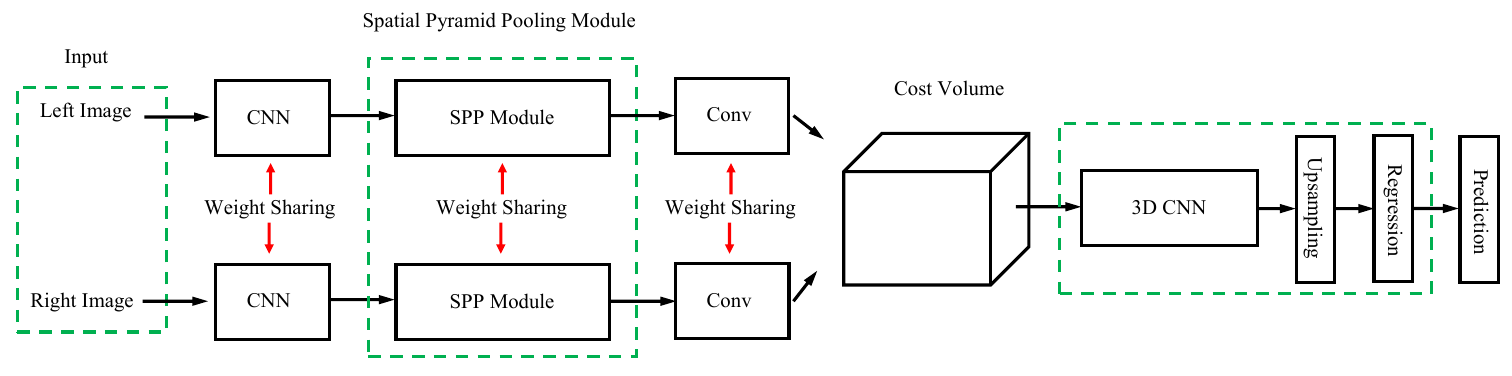}
    \caption{The architecture of Pyramid Stereo Matching Network (PSMNet)}
    \label{fig_PSMNet}
\end{figure*}

To perform the depth detection, information on the object is needed. RetinaNet ant colony detection conducts target recognition. For depth detection, the stereo images are used to determine the distance from the camera, which can be installed in the drone’s center of gravity. In this regard, PSMNet is utilized to provide depth estimation from a stereo pair of images \cite{chang2018pyramid}. PSMNet utilizes global information in stereo matching using spatial pyramid pooling \cite{he2015spatial,zhao2017pyramid} and dilated convolution \cite{chen2017deeplab}. The architecture of PSMNet is illustrated in Fig. \ref{fig_PSMNet}.

Several left-right images are utilized by a stereo disparity estimation algorithm, from which they are captured by two cameras with a horizontal offset (i.e., baseline ($b$)). The output of disparity estimation ($Y$) is the same as either the left or right images. Generally, the depth estimation algorithm uses the left image as a reference and records in Y; thereby, the horizontal disparity is applied to the right image for each pixel. 
Together with the horizontal focal length $f_u$ of the left camera, the depth map $D$ is derived as follows:

\begin{equation}\label{DroFac_eq43}
D(u,v) = \frac{f_u \times b}{Y(u,v)}
\end{equation}

\begin{figure}[h]
    \centering
    \includegraphics[scale = 0.7]{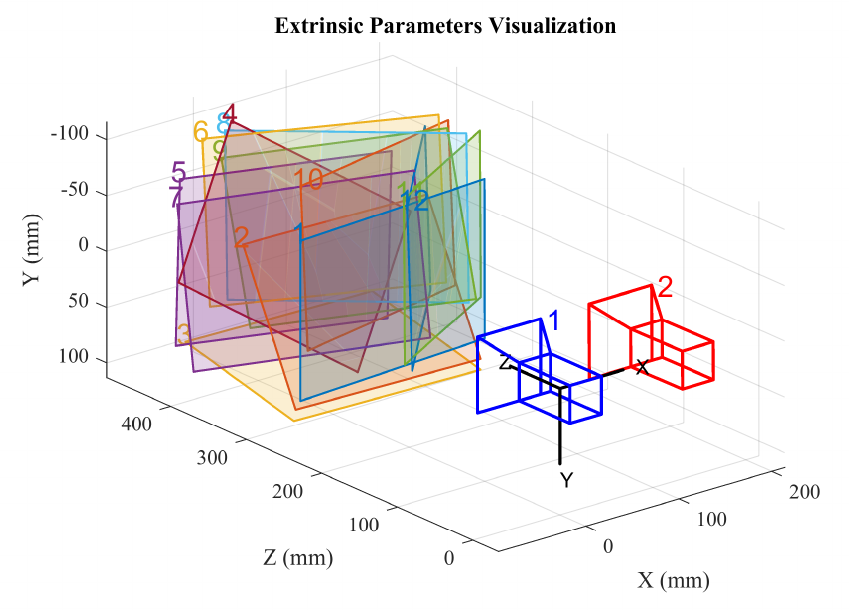}
    \caption{Visualization of Extrinsic parameters}
    \label{fig_CamPar}
\end{figure}

Thus, the 3D location $(x,y,z)$ of each pixel $(u,v)$ of the target, which can be used to calculate the relative distance between drone and target, is formulated as below:

\begin{equation}\label{DroFac_eq44}
z = D(u,v)
\end{equation}

\begin{equation}\label{DroFac_eq45}
x = \frac{(u-c_u)\times z}{f_u}
\end{equation}

\begin{equation}\label{DroFac_eq46}
y = \frac{(v-c_v)\times z}{f_v}
\end{equation}
where $(c_u,c_v)$ is the pixel location corresponding to the camera center and $f_v$ is the vertical focal length. Thus, the extrinsic parameters of simulated cameras with $150~mm$ offset can be visualized in Fig. \ref{fig_CamPar}.

\begin{figure}
    \centering
    \includegraphics[scale = 0.8]{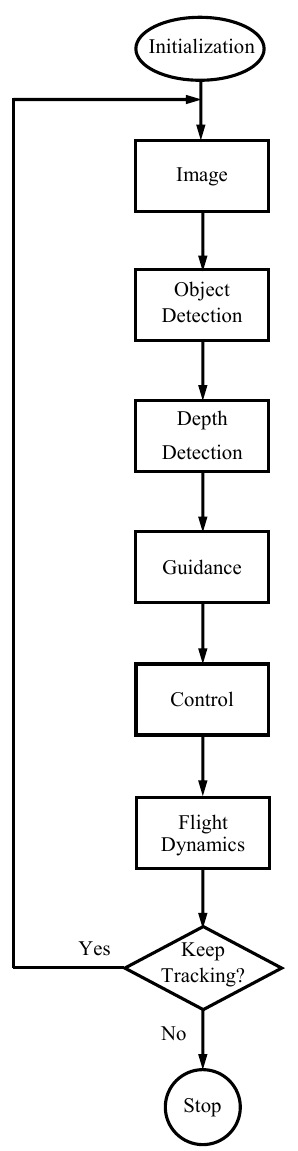}
    \caption{The general flowchart of the proposed system}
    \label{fig_generalChart}
\end{figure}

The guidance module is needed to utilize object and depth detection. The proposed independent drone uses the image as an input for autonomous flying. In this regard, the image as input passes through the object detection to provide the object information for the depth detection modules. Depth detection estimates the relative distance from the target. Further, the guidance law (PPN) generates the commands for the drone to reach the target by applying the proposed PIDA controller. Finally, the controller takes action on flight dynamics. The general flowchart of the proposed system is presented in Fig. \ref{fig_generalChart}. 

\section{Numerical Results} \label{DroFac_sec8}

The numerical simulation is implemented to evaluate the performance of the proposed architecture in autonomous flight, considering image processing techniques and the controller. The model quadcopter was simulated in MATLAB R2018b in a Simulink environment on Windows 10 with an Intel(R) Core(TM)i7-6700 CPU @ 3.4 Hz. The quadcopter parameters are listed in Table \ref{DroFac_table2}.
 
\begin{table}[h]
\centering
\caption{Quadcopter Model Parameters}
\label{DroFac_table2}
\begin{tabular}{c p{3.5cm} c c}
\hline
 
\textbf{Parameter} & \textbf{Description} & \textbf{value} & \textbf{Unit} \\ 

\hline

$m$ & Mass & $0.8$ & kg \\
$l$ & Arm length & $0.2$ & m \\
$g$ & Gravity acceleration & $9.81$ & $m/s^2$ \\
$c$ & Force to torque coefficient & $3e-5$ & kg $m^2$ \\
$I_{xx}$ & Body moment of inertia along x-axis & $2.28e-2$ &  kg $m^2$\\
$I_{yy}$ & Body moment of inertia along y-axis& $3.10e-2$ & kg $m^2$\\
$I_{zz}$ & Body moment of inertia along z-axis& $4.40e-2$ & kg $m^2$\\
$I_{m}$  & Motor moment of inertia & $8.3e-5$ & kg $m^2$\\

\hline
\end{tabular}
\end{table}

\begin{figure}
 \centering
 
\begin{subfigure}
		{\includegraphics[scale = 0.7]{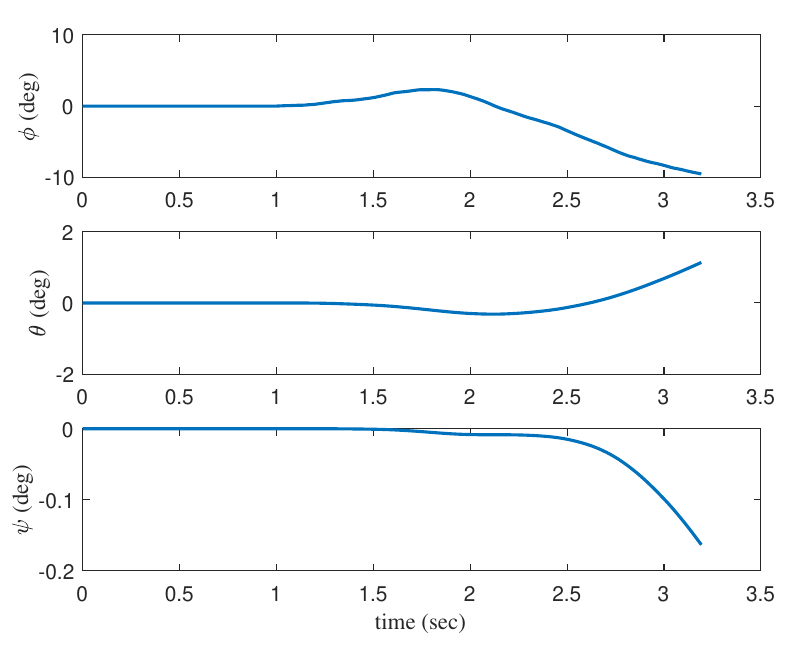}}
	\caption{Roll, Pitch and Yaw angle in noisy environment without controller}
	\label{DroFac_fig55}
\end{subfigure}
		
		\quad
		
	\begin{subfigure}
		
    {\includegraphics[scale = 0.7]{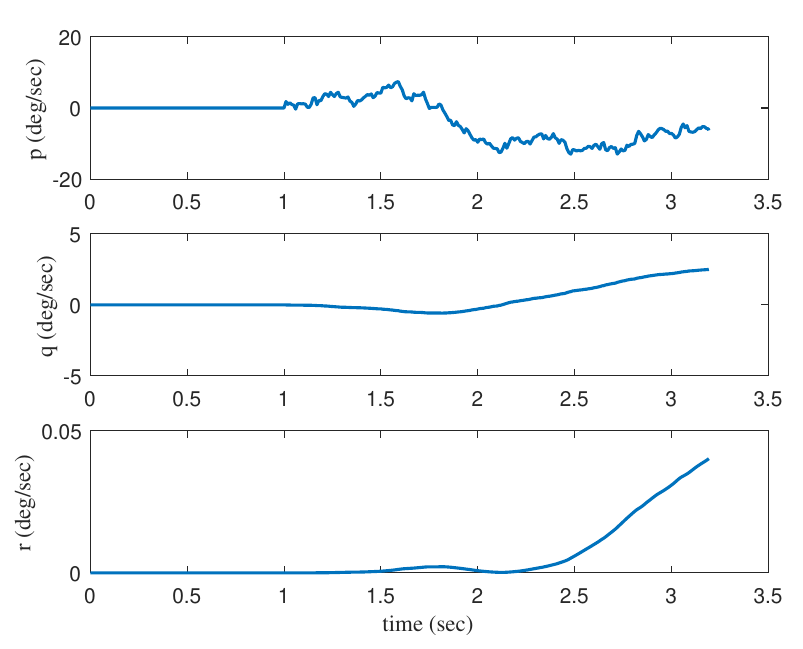}}
		\caption{Angular velocity of the modeled drone in noisy environment without controller}
		\label{DroFac_fig66}
	\end{subfigure}
	
\end{figure}

To begin the simulation and tune the hyper-parameters, the initial state is introduced to identify the best optimal parameters. In this study, it is assumed that the initial altitude and velocity are $X_E = [0~0~ $-50$]^T~m$ and $V = [u~v~w]^T = [1~1~0]^T~m/s$, respectively. A disturbance is applied to the quadcopter and is modeled as white noise (mean value ($\mu$) is zero and standard deviation ($\sigma$) is one) at time $1~sec$ in the roll channel. This disturbance destabilizes the system and locates the eigenvalues of $A$ in the right half-plane (see Fig. \ref{DroFac_fig55} and Fig. \ref{DroFac_fig66}). Additionally, the quadrotor is highly sensitive to the noisy environment because of instability and cross-coupling. In this regard, PIDA with derivative filter, which obviates the noise from the measurement inputs, is designed to respond to this issue and keep the flight stable. Note that the hyper-parameters of the proposed system are tuned according to the model.

\begin{figure}[h]
    \centering
    \includegraphics[scale = 0.7]{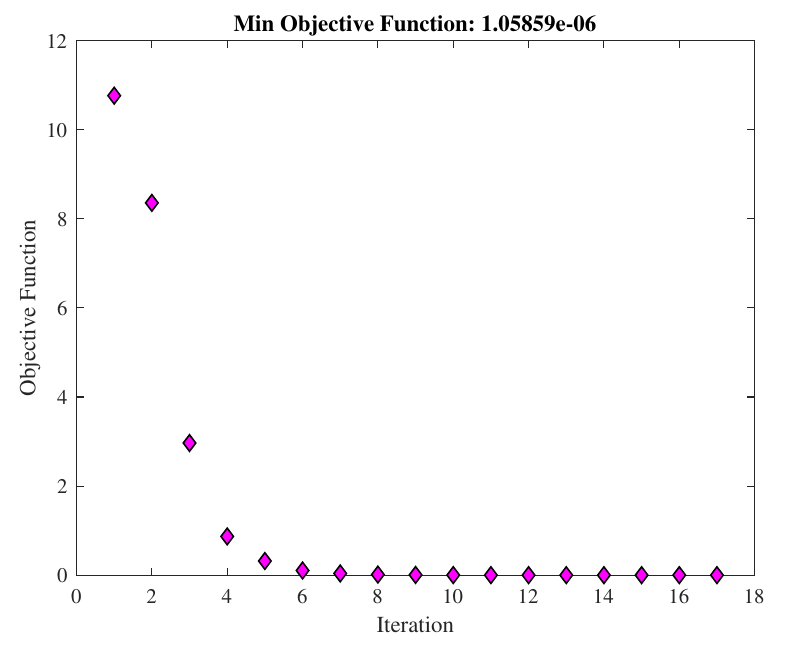}
    \caption{Performance of SDSA versus iteration }
    \label{DroFac_fig1&1}
\end{figure}

According to the proposed PIDA with a derivative filter, tracking desire inputs, which can be defined as commands to the quadcopter, are another issue that can be addressed by a MIMO controller (i.e., four inputs and four outputs). The proposed controller can be set by four gains and the time constant for each mode/channel. The controllers’ parameters are tuned using SDSA \cite{ZandaviSDSA2019}, convergence graph of which is shown in Fig. \ref{DroFac_fig1&1}. The SDSA is applied to the objective function introduced in Eq (\ref{DroFac_eq222}). Table \ref{DroFac_table3} lists the outputs of the heuristic optimization algorithm as the best fit set of parameters for different modes/channels.

\begin{table}
\centering
\caption{Controller Parameters for Altitude and Attitude}
\label{DroFac_table3}
\begin{tabular}{lcccc}
\hline

\textbf{Controller Parameter} & \multicolumn{4}{c}{\textbf{Channel}}\\

\cline{2-5}

& Roll & Pitch & Yaw & Altitude \\

\hline

$k_i$ & $0.1436$ & $3.6869$ & $0.0437$ & $1.00$\\
$k_d$ & $6.5097$ & $21.2743$& $29.9872$& $11.4676$ \\
$k_a$ & $0.5772$ & $0.3429$ & $23.5238$& $7.5114$ \\
$T_f$ & $0.0437$ & $0.0331$ & $0.0117$ & $0.3752$ \\








\hline
\end{tabular}
\end{table}

\begin{figure}[h]
    \centering
    \includegraphics[scale = 0.7]{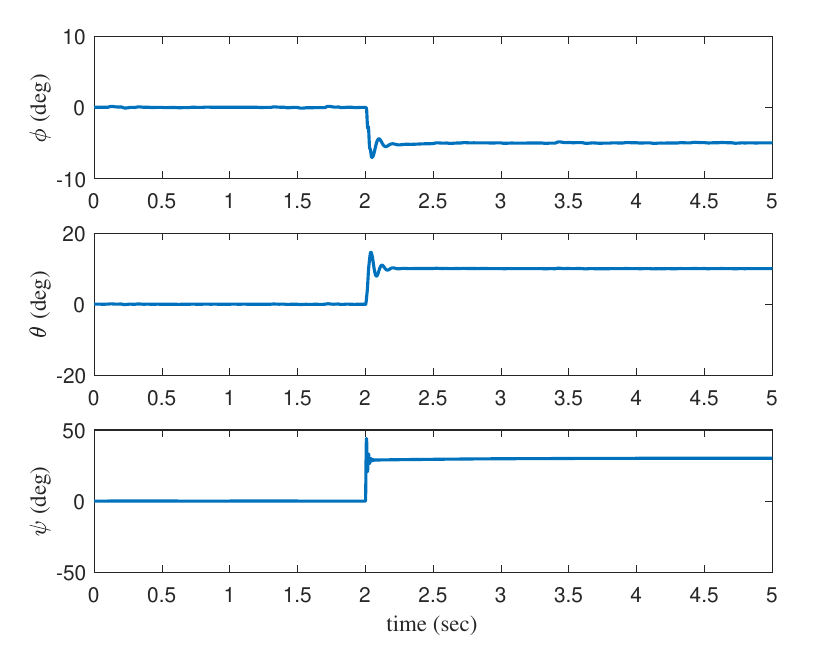}
    \caption{Controller response to the change of Roll, Pitch and Yaw angle}
    \label{DroFac_fig2}
\end{figure}

\begin{figure}[h]
    \centering
    \includegraphics[scale = 0.7]{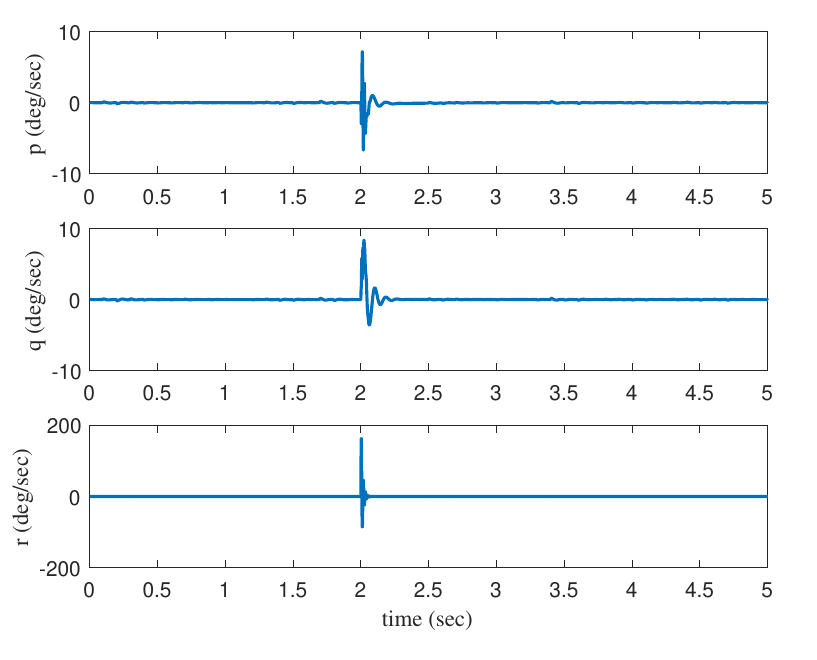}
    \caption{Controller response to the change of Angular velocity}
    \label{DroFac_fig3}
\end{figure}

\begin{figure}[h]
    \centering
    \includegraphics[scale = 0.7]{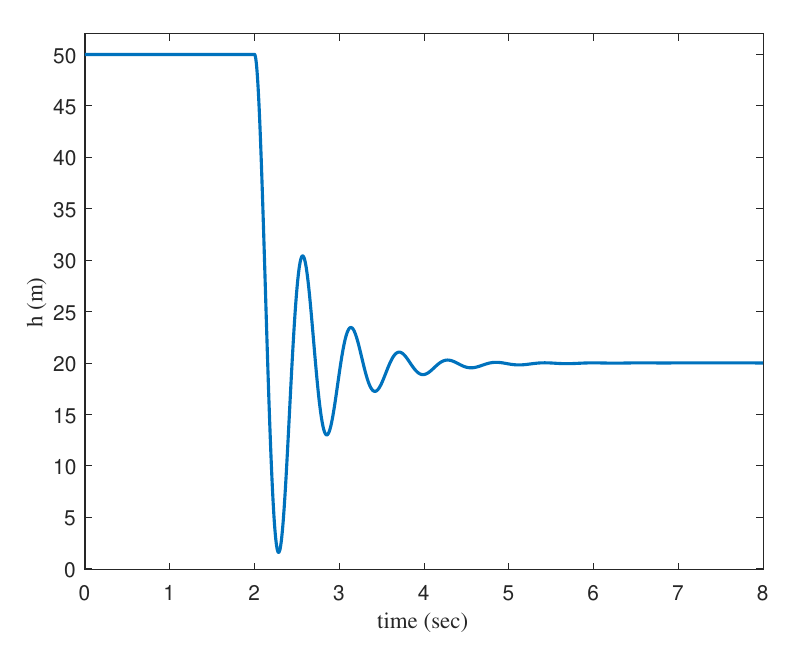}
    \caption{Controller response to the change of Altitude}
    \label{DroFac_fig4}
\end{figure}

The complex commands that enable coupling among different modes of the modeled quadcopter are used to evaluate the performance of the designed controller. New command angles are provided by a step function with $2~sec$ delay time in the simulation environment, where $\phi = -5^{\circ}$, $\theta = 10 ^{\circ}$, $\psi = 30 ^{\circ}$ and with altitude starting from $50~m$ and stabling at $20~m$. Note that noisy measurements have been considered for this simulation, and are modeled as white noise. As the simulation results demonstrate, the noise cannot affect the performance of the quadcopter. Figures \ref{DroFac_fig2}–\ref{DroFac_fig4} show that the proposed controller can properly respond to and track the reference commands in the noisy environment.

 

		
	
	
		
	

\begin{figure}
    \centering
				
		\includegraphics[width=0.45\textwidth,trim ={3cm 8.6cm 1cm 10cm},clip,scale = 0.6]{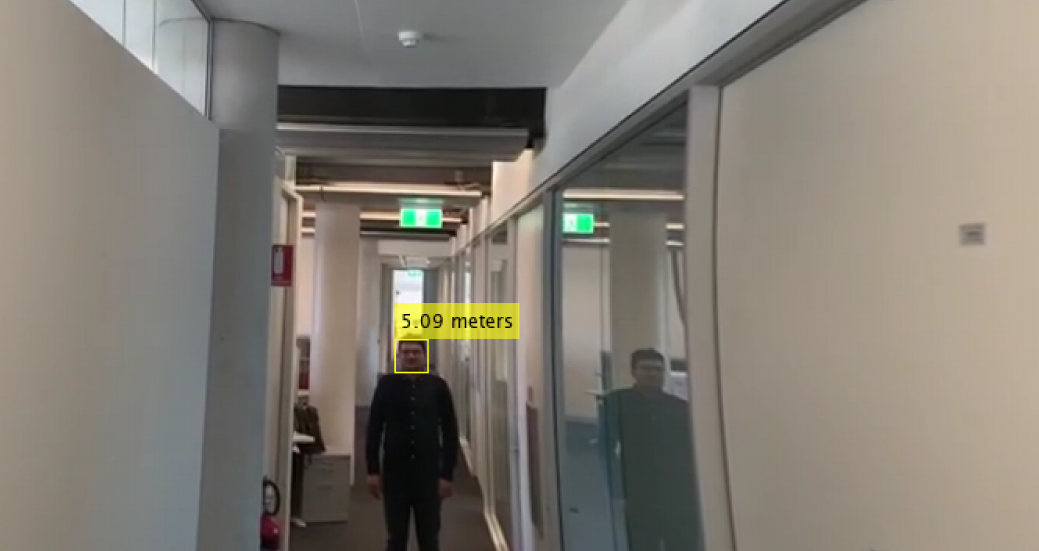} \quad
		\includegraphics[width=0.45\textwidth,trim ={3cm 8.6cm 1cm 10cm},clip,scale = 0.6]{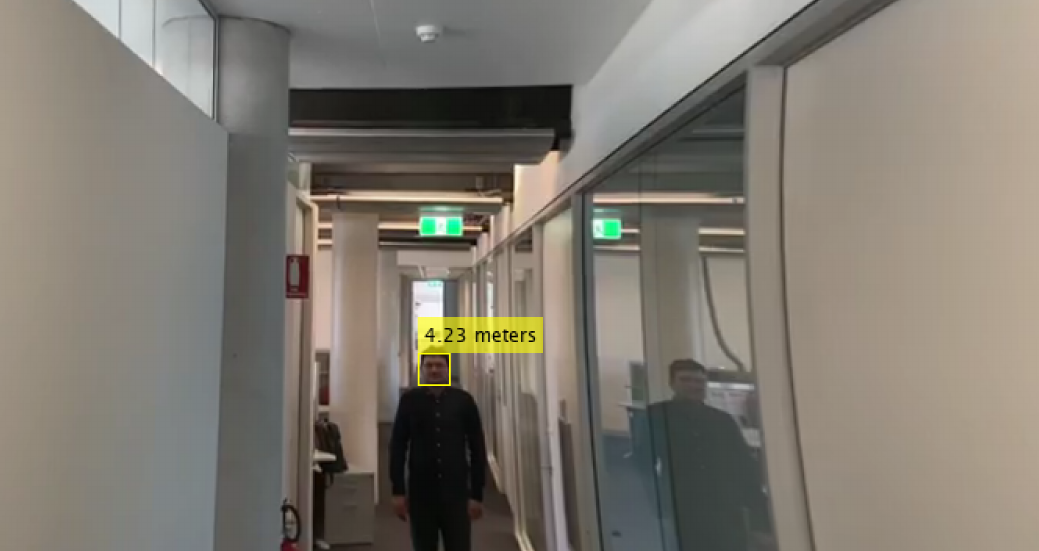}
		
		
		\includegraphics[width=0.45\textwidth,trim ={3cm 8.6cm 1cm 10cm},clip,scale = 0.6]{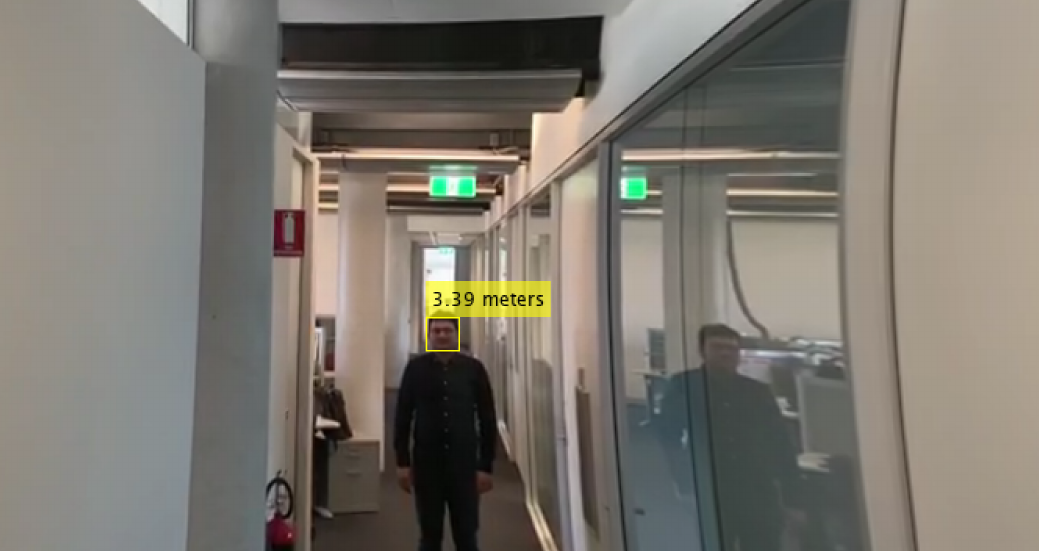} \quad
		\includegraphics[width=0.45\textwidth,trim ={3cm 8.6cm 1cm 10cm},clip,scale = 0.6]{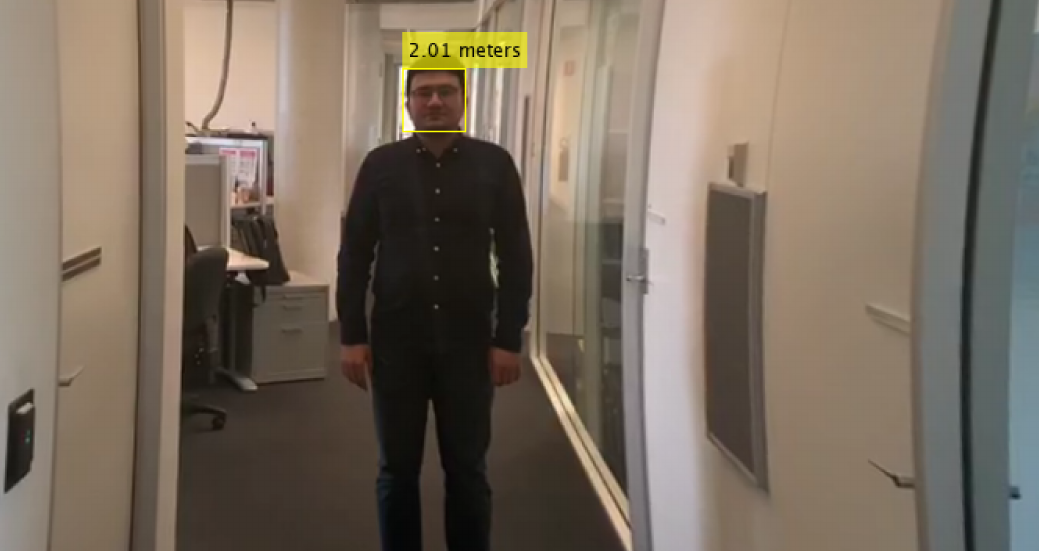}
		
		
						
		\caption{Object and Depth Detection of the target}
		
		\label{fig_FirstLastFrame}
		
\end{figure}

Table \ref{DroFac_table3} shows the optimal parameters that are tuned by SDSA. From the experiments, a scenario is defined to evaluate the workflow of the system: a target is a stationary person located at $X_{E_T} = [5~5~0]~m$ in an ECI frame that is considered a local frame. The drone is simulated in an indoor environment with the initial position at $X_{E_D} = [0~0~-5]~m$. A mission is used to instruct the drone to reach the target by maintaining the safe distance (i.e., $2~m$ is considered a suitable threshold from the object). Thus, RetinaNet ant colony detection and PSMNet recognize the relative distance to the target via the camera and the command is readily enacted. For example, the object and depth detection of the target in four different sampling times are shown in Fig. \ref{fig_FirstLastFrame}.

\begin{figure}
    \centering
    \includegraphics[scale = 0.5]{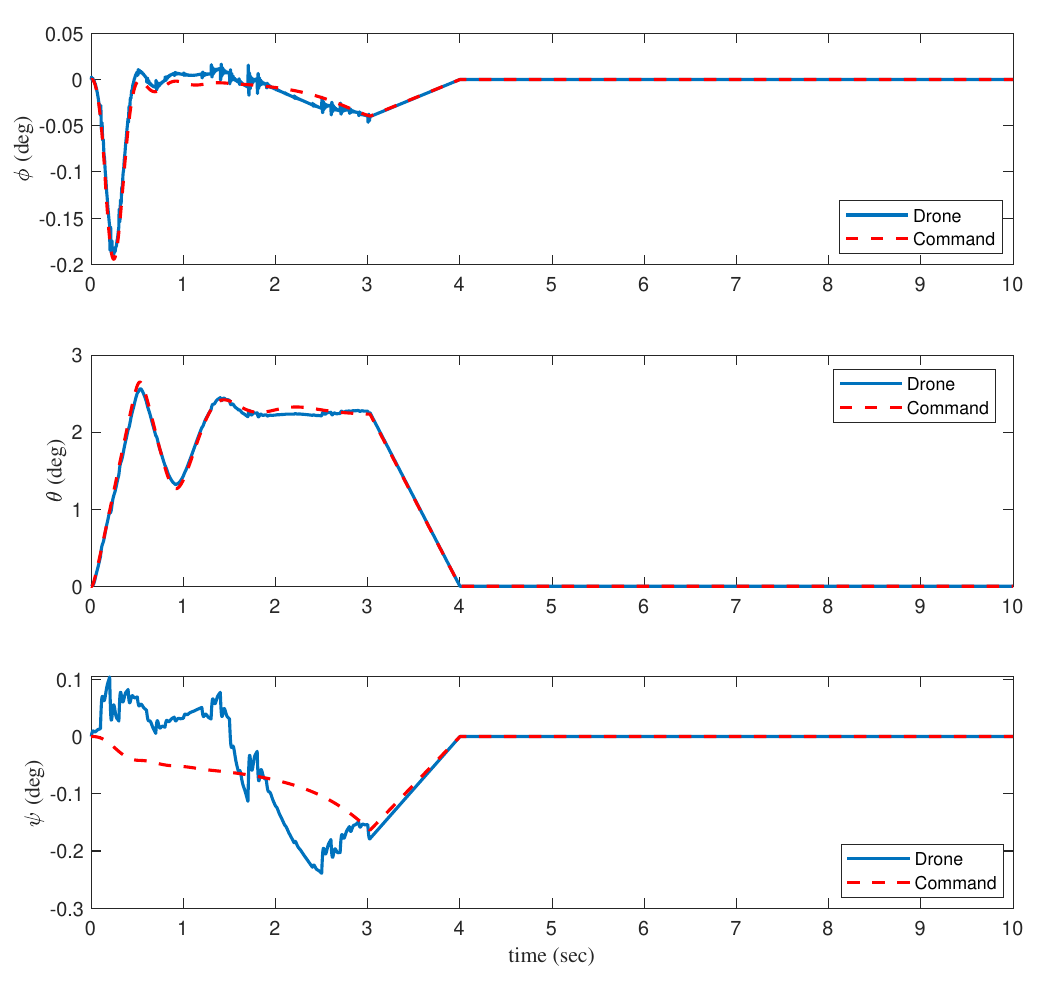}
    \caption{Controller response to the commands in different channels}
    \label{fig_PTSCommand}
\end{figure}

\begin{figure}
    \centering
    \includegraphics[scale = 0.7]{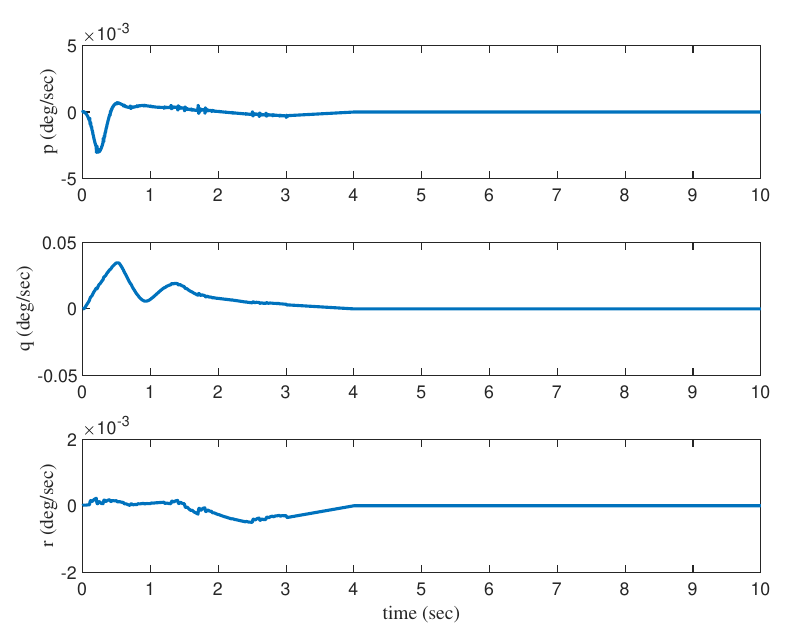}
    \caption{Controller response to the commands in different channels}
    \label{fig_PQRcommand}
\end{figure}

\begin{figure}
    \centering
    \includegraphics[scale = 0.7]{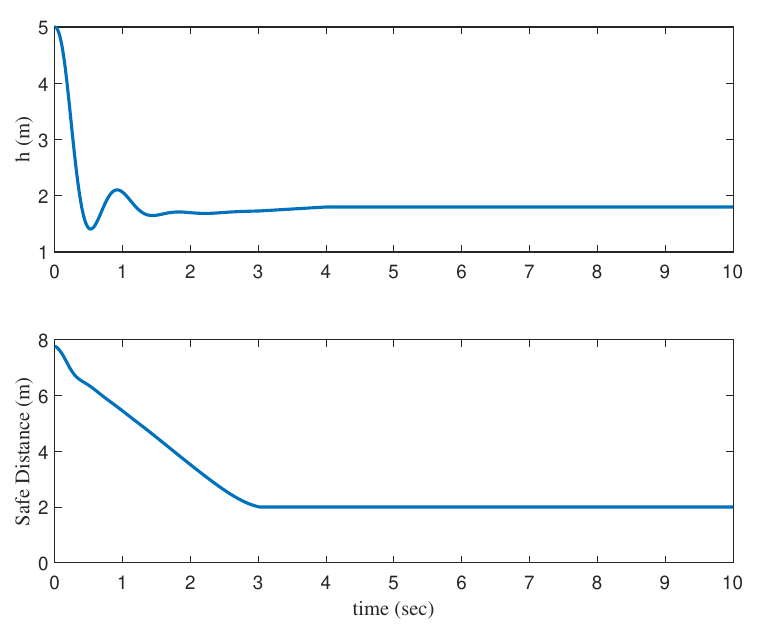}
    \caption{Height and safe distance of the drone over time}
    \label{fig_SafeDistance}
\end{figure}

\begin{figure}
    \centering
    \includegraphics[scale = 0.7]{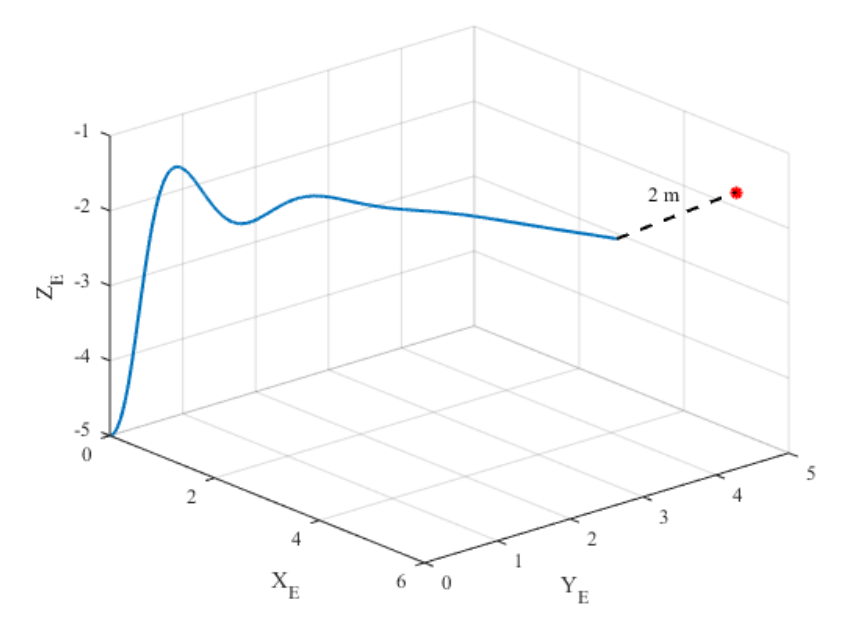}
    \caption{Drone Trajectory}
    \label{fig_DroneTrajectory}
\end{figure}

Simultaneously, the relative distance calculated by the image processing module is utilized by the guidance discipline and followed by control and flight dynamics systems. The simulation results show that the proposed system is adept at tracking the target in consideration of the noisy environment. The control responses and trajectory of the drone are illustrated in Fig. \ref{fig_PTSCommand}$–$\ref{fig_DroneTrajectory}. As shown, the controller is tracking the desire input generated by the guidance law over time. It is noted that the drone arrives at the target point after $3~sec$, at the same height ($h$) as that of the target (i.e., $h = 1.8~m$). Further, the drone stays in its position to meet the safe threshold requirement (i.e., safe distance $= 2~m$ ). Fig. \ref{fig_PTSCommand} and Fig. \ref{fig_PQRcommand} demonstrate that the quadcopter moves smoothly to touch the target because angular velocity fluctuates minimally around zero, and the Euler angles converge on zero to maintain both the height and safe distance to stabilize and approach around $2~m$ (see Fig. \ref{fig_SafeDistance}). 
 
\section{Conclusion} \label{DroFac_sec9}

This paper has proposed a new workflow to use images as inputs for the controller to achieve autonomous flight while considering the noisy indoor environment and uncertainties. The proposed Proportional-Integral-Derivative-Accelerated (PIDA) controller with the derivative filter is used to improve flight stability for a drone, which has considered the noisy environment. The paper has also proposed a platform to adapt deep learning-based object and depth detection techniques to fly the drone autonomously in the indoor environment surrounded by uncertainties. The mathematical model considering non-linearity, uncertainties, and coupling was derived from an accurate model with a high level of fidelity.The simulation results show that image processing techniques (RetinaNet ant colony detection and PSMNet) and the proposed PIDA controller tuned by tochastic Dual Simplex Algorithm (SDSA) are able to track the desired point in the presence of disturbances. 



\bibliographystyle{IEEEtran}

\bibliography{bibliofile} 

\begin{thebibliography}{10}
\providecommand{\url}[1]{#1}
\csname url@samestyle\endcsname
\providecommand{\newblock}{\relax}
\providecommand{\bibinfo}[2]{#2}
\providecommand{\BIBentrySTDinterwordspacing}{\spaceskip=0pt\relax}
\providecommand{\BIBentryALTinterwordstretchfactor}{4}
\providecommand{\BIBentryALTinterwordspacing}{\spaceskip=\fontdimen2\font plus
\BIBentryALTinterwordstretchfactor\fontdimen3\font minus
  \fontdimen4\font\relax}
\providecommand{\BIBforeignlanguage}[2]{{%
\expandafter\ifx\csname l@#1\endcsname\relax
\typeout{** WARNING: IEEEtran.bst: No hyphenation pattern has been}%
\typeout{** loaded for the language `#1'. Using the pattern for}%
\typeout{** the default language instead.}%
\else
\language=\csname l@#1\endcsname
\fi
#2}}
\providecommand{\BIBdecl}{\relax}
\BIBdecl

\bibitem{kim2019adaptive}
S.-K. Kim and C.~K. Ahn, ``Adaptive nonlinear tracking control algorithm for
  quadcopter applications,'' \emph{IEEE Transactions on Aerospace and
  Electronic Systems}, 2019.

\bibitem{koh2012dawn}
L.~P. Koh and S.~A. Wich, ``Dawn of drone ecology: low-cost autonomous aerial
  vehicles for conservation,'' \emph{Tropical Conservation Science}, vol.~5,
  no.~2, pp. 121--132, 2012.

\bibitem{phung2017enhanced}
M.~D. Phung, C.~H. Quach, T.~H. Dinh, and Q.~Ha, ``Enhanced discrete particle
  swarm optimization path planning for uav vision-based surface inspection,''
  \emph{Automation in Construction}, vol.~81, pp. 25--33, 2017.

\bibitem{rajappa2016adaptive}
S.~Rajappa, C.~Masone, H.~H. B{\"u}lthoff, and P.~Stegagno, ``Adaptive super
  twisting controller for a quadrotor uav,'' in \emph{2016 IEEE International
  Conference on Robotics and Automation (ICRA)}.\hskip 1em plus 0.5em minus
  0.4em\relax IEEE, 2016, pp. 2971--2977.

\bibitem{derafa2012super}
L.~Derafa, A.~Benallegue, and L.~Fridman, ``Super twisting control algorithm
  for the attitude tracking of a four rotors uav,'' \emph{Journal of the
  Franklin Institute}, vol. 349, no.~2, pp. 685--699, 2012.

\bibitem{zuo2010trajectory}
Z.~Zuo, ``Trajectory tracking control design with command-filtered compensation
  for a quadrotor,'' \emph{IET control theory \& applications}, vol.~4, no.~11,
  pp. 2343--2355, 2010.

\bibitem{raffo2010integral}
G.~V. Raffo, M.~G. Ortega, and F.~R. Rubio, ``An integral predictive/nonlinear
  h∞ control structure for a quadrotor helicopter,'' \emph{Automatica},
  vol.~46, no.~1, pp. 29--39, 2010.

\bibitem{ritz2011quadrocopter}
R.~Ritz, M.~Hehn, S.~Lupashin, and R.~D'Andrea, ``Quadrocopter performance
  benchmarking using optimal control,'' in \emph{2011 IEEE/RSJ International
  Conference on Intelligent Robots and Systems}.\hskip 1em plus 0.5em minus
  0.4em\relax IEEE, 2011, pp. 5179--5186.

\bibitem{zandavi2018multidisciplinary}
S.~M. Zandavi and S.~H. Pourtakdoust, ``Multidisciplinary design of a guided
  flying vehicle using simplex nondominated sorting genetic algorithm ii,''
  \emph{Structural and Multidisciplinary Optimization}, vol.~57, no.~2, pp.
  705--720, 2018.

\bibitem{xu2006sliding}
R.~Xu and U.~Ozguner, ``Sliding mode control of a quadrotor helicopter,'' in
  \emph{Proceedings of the 45th IEEE Conference on Decision and Control}.\hskip
  1em plus 0.5em minus 0.4em\relax IEEE, 2006, pp. 4957--4962.

\bibitem{besnard2007control}
L.~Besnard, Y.~B. Shtessel, and B.~Landrum, ``Control of a quadrotor vehicle
  using sliding mode disturbance observer,'' in \emph{2007 American Control
  Conference}.\hskip 1em plus 0.5em minus 0.4em\relax IEEE, 2007, pp.
  5230--5235.

\bibitem{ang2005pid}
K.~H. Ang, G.~Chong, and Y.~Li, ``Pid control system analysis, design, and
  technology,'' \emph{IEEE transactions on control systems technology},
  vol.~13, no.~4, pp. 559--576, 2005.

\bibitem{jung1996analytic}
S.~Jung and R.~C. Dorf, ``Analytic pida controller design technique for a third
  order system,'' in \emph{Proceedings of 35th IEEE Conference on Decision and
  Control}, vol.~3.\hskip 1em plus 0.5em minus 0.4em\relax IEEE, 1996, pp.
  2513--2518.

\bibitem{zandavi2017novel}
S.~M. Zandavi, F.~Sha, V.~Chung, Z.~Lu, and W.~Zhi, ``A novel ant colony
  detection using multi-region histogram for object tracking,'' in
  \emph{International Conference on Neural Information Processing}.\hskip 1em
  plus 0.5em minus 0.4em\relax Springer, 2017, pp. 25--33.

\bibitem{lin2017focal}
T.-Y. Lin, P.~Goyal, R.~Girshick, K.~He, and P.~Doll{\'a}r, ``Focal loss for
  dense object detection,'' in \emph{Proceedings of the IEEE international
  conference on computer vision}, 2017, pp. 2980--2988.

\bibitem{chang2018pyramid}
J.-R. Chang and Y.-S. Chen, ``Pyramid stereo matching network,'' in
  \emph{Proceedings of the IEEE Conference on Computer Vision and Pattern
  Recognition}, 2018, pp. 5410--5418.

\bibitem{ZandaviSDSA2019}
S.~M. {Zandavi}, V.~Y.~Y. {Chung}, and A.~{Anaissi}, ``Stochastic dual simplex
  algorithm: A novel heuristic optimization algorithm,'' \emph{IEEE
  Transactions on Cybernetics}, pp. 1--10, 2019.

\bibitem{zipfel2007modeling}
P.~H. Zipfel, \emph{Modeling and simulation of aerospace vehicle
  dynamics}.\hskip 1em plus 0.5em minus 0.4em\relax American Institute of
  Aeronautics and Astronautics, 2007.

\bibitem{rao2009engineering}
S.~S. Rao and S.~S. Rao, \emph{Engineering optimization: theory and
  practice}.\hskip 1em plus 0.5em minus 0.4em\relax John Wiley \& Sons, 2009.

\bibitem{siouris2004missile}
G.~M. Siouris, \emph{Missile guidance and control systems}.\hskip 1em plus
  0.5em minus 0.4em\relax Springer Science \& Business Media, 2004.

\bibitem{nobahari2016simplex}
H.~Nobahari, S.~M. Zandavi, and H.~Mohammadkarimi, ``Simplex filter: a novel
  heuristic filter for nonlinear systems state estimation,'' \emph{Applied Soft
  Computing}, vol.~49, pp. 474--484, 2016.

\bibitem{zandavi2019state}
S.~M. Zandavi and V.~Chung, ``State estimation of nonlinear dynamic system
  using novel heuristic filter based on genetic algorithm,'' \emph{Soft
  Computing}, vol.~23, no.~14, pp. 5559--5570, 2019.

\bibitem{yu2016unitbox}
J.~Yu, Y.~Jiang, Z.~Wang, Z.~Cao, and T.~Huang, ``Unitbox: An advanced object
  detection network,'' in \emph{Proceedings of the 24th ACM international
  conference on Multimedia}.\hskip 1em plus 0.5em minus 0.4em\relax ACM, 2016,
  pp. 516--520.

\bibitem{girshick2015fast}
R.~Girshick, ``Fast r-cnn,'' in \emph{Proceedings of the IEEE international
  conference on computer vision}, 2015, pp. 1440--1448.

\bibitem{chi2019selective}
C.~Chi, S.~Zhang, J.~Xing, Z.~Lei, S.~Z. Li, and X.~Zou, ``Selective refinement
  network for high performance face detection,'' in \emph{Proceedings of the
  AAAI Conference on Artificial Intelligence}, vol.~33, 2019, pp. 8231--8238.

\bibitem{he2015spatial}
K.~He, X.~Zhang, S.~Ren, and J.~Sun, ``Spatial pyramid pooling in deep
  convolutional networks for visual recognition,'' \emph{IEEE transactions on
  pattern analysis and machine intelligence}, vol.~37, no.~9, pp. 1904--1916,
  2015.

\bibitem{zhao2017pyramid}
H.~Zhao, J.~Shi, X.~Qi, X.~Wang, and J.~Jia, ``Pyramid scene parsing network,''
  in \emph{Proceedings of the IEEE conference on computer vision and pattern
  recognition}, 2017, pp. 2881--2890.

\bibitem{chen2017deeplab}
L.-C. Chen, G.~Papandreou, I.~Kokkinos, K.~Murphy, and A.~L. Yuille, ``Deeplab:
  Semantic image segmentation with deep convolutional nets, atrous convolution,
  and fully connected crfs,'' \emph{IEEE transactions on pattern analysis and
  machine intelligence}, vol.~40, no.~4, pp. 834--848, 2017.

\end{thebibliography}


\end{document}